\documentclass[10pt,twocolumn,letterpaper]{article}

\usepackage{cvpr}              \usepackage{bbding}

\usepackage[dvipsnames]{xcolor}

\definecolor{cvprblue}{rgb}{0.21,0.49,0.74}
\usepackage[pagebackref,breaklinks,colorlinks,citecolor=cvprblue]
{hyperref}

\title{A Stitch in Time: Learning Procedural Workflow via Self-Supervised Plackett–Luce Ranking}

\author{Chengan Che, Chao Wang, Xinyue Chen, Sophia Tsoka, Luis C. Garcia-Peraza-Herrera \\
Department of Informatics, King's College London, UK \\
{\tt\small \{chengan.che, chao.wang, xinyue.1.chen, sophia.tsoka, luis\_c.garcia\_peraza\_herrera\}@kcl.ac.uk}
}

\usepackage{booktabs}
\usepackage{graphicx}
\usepackage{amsmath}
\usepackage{multirow} \usepackage{tabularx}  \usepackage{listings}
\usepackage{hyperref}
\usepackage{subcaption}
\usepackage{pifont} 

\usepackage[table]{xcolor}

\usepackage{xcolor}
\usepackage{algorithm}
\usepackage{algpseudocode}

\definecolor{MyGreen}{HTML}{E2EFDA}      \definecolor{MyBlue}{HTML}{DDEBF7} 
\definecolor{lightblue}{HTML}{75C6FF} 
\definecolor{videoblue}{HTML}{4E85D9} \definecolor{imageorange}{HTML}{E97133} \usepackage{placeins}

\begin{document}
\maketitle
\begin{abstract}

Procedural activities, ranging from routine cooking to complex surgical operations, are highly structured sequences of actions performed in a specific temporal order. 
Despite the success of current self-supervised learning (SSL) methods on static images and short clips, these models often overlook the underlying sequential structure of such activities. 
We expose this lack of procedural awareness with a motivating experiment: models pretrained on forward and time-reversed sequences produce highly similar features, confirming that their representations are blind to the underlying procedural order.
To address this shortcoming, we propose \textbf{PL-Stitch}, a self-supervised framework that harnesses the inherent temporal order of video frames as a powerful supervisory signal.
Our approach integrates two novel probabilistic objectives based on the Plackett-Luce (PL) model.
The primary PL objective trains the model to sort sampled frames chronologically, compelling it to learn the global workflow progression. 
The secondary objective, a spatio-temporal jigsaw loss, complements the learning by capturing fine-grained, cross-frame object correspondences.
Our approach consistently achieves superior performance across five surgical and cooking benchmarks.
Specifically, PL-Stitch yields significant gains in surgical phase recognition (e.g., +11.4\,pp in k-NN accuracy on Cholec80) and cooking action segmentation (e.g., +5.7\,pp in linear probing accuracy on Breakfast), demonstrating its effectiveness for procedural video representation learning.
Code and models are available at \href{https://github.com/visurg-ai/PL-Stitch}{https://github.com/visurg-ai/PL-Stitch}.

\begin{figure}[!t]
    \centering 

\begin{subfigure}{0.99\columnwidth}
        \centering
\includegraphics[width=\linewidth]{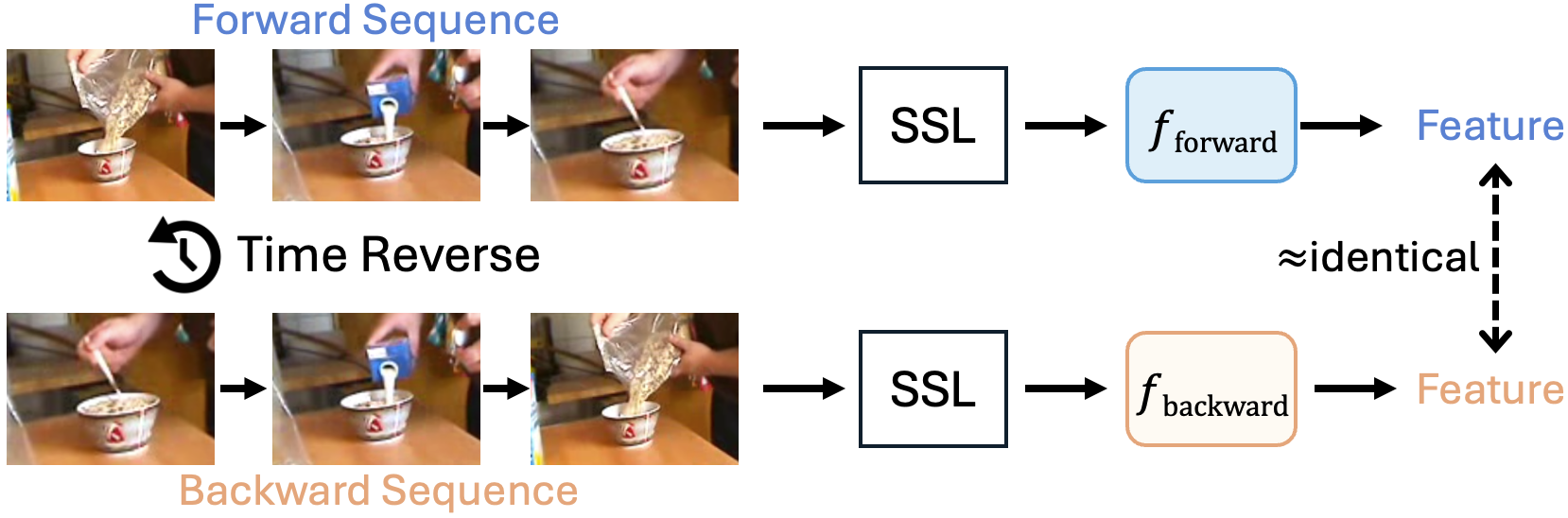}
        \caption{Procedural agnosticism of prior SSL methods}
        \label{fig:figure1a}
    \end{subfigure}

    \vspace{0.2cm}

\begin{subfigure}{0.99\columnwidth}
        \centering
        \includegraphics[width=\linewidth]{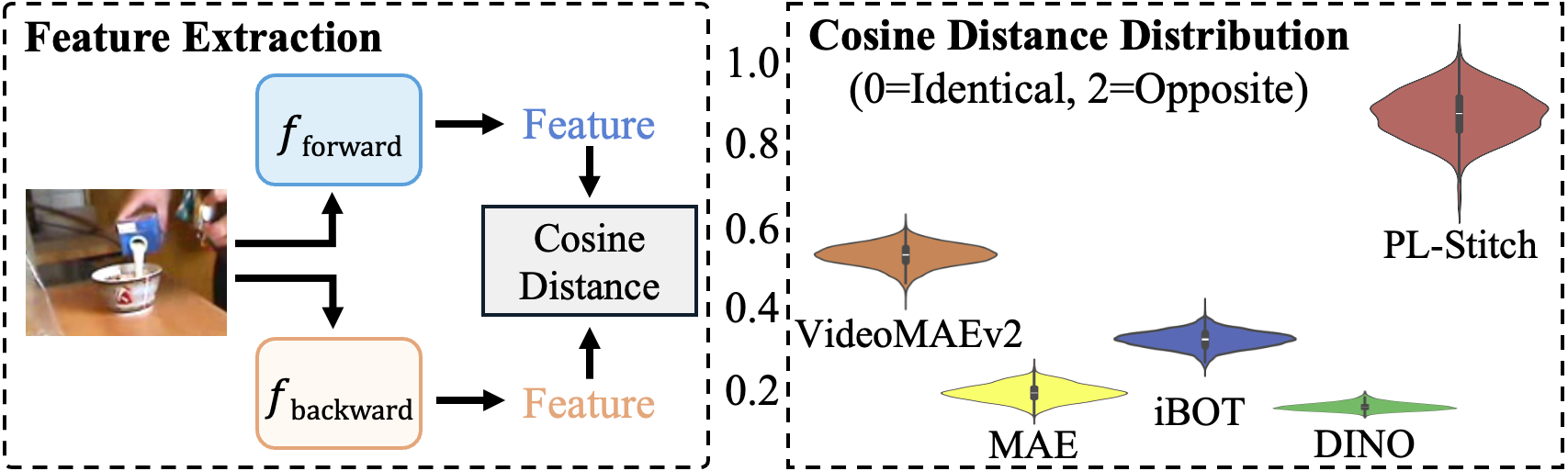}
        \caption{Experimental validation of procedural awareness}
        \label{fig:figure1b}
    \end{subfigure}

\caption{
        \textbf{Procedural awareness in self-supervised learning (SSL).}
(a) Prior SSL methods are \textit{procedural-agnostic}, learning features that fail to capture the procedural structure.
(b) We pretrain PL-Stitch and prior SSL methods on the Breakfast dataset~\cite{Kuehne2014TheActivities} using \textit{forward} and \textit{backward} sequences.
The violin plot shows the cosine distance between the feature vectors of the same video frames. Baselines exhibit procedural agnosticism (low distance), whereas PL-Stitch's high distance validates its procedural awareness.
    }
    \label{fig:figure1}
\end{figure}

\end{abstract}

 \section{Introduction}
\label{sec:intro}

Many human activities, from daily cooking to surgical operations, are defined by a procedural workflow, a sequence of multi-step actions in a specific temporal order. 
This raises a critical question: do current self-supervised learning (SSL) methods actually learn this procedural logic, or do they merely recognize static actions? 

To provide insights on this question, we conducted an experiment that motivates our work. We pretrained leading SSL models on both forward (chronological) and backward (time-reversed) video sequences of the Breakfast dataset~\cite{Kuehne2014TheActivities}.
At inference, we feed the same video frames to both the forward and backward trained encoders and compute the cosine distance between their resulting features. 
As shown in Fig.~\ref{fig:figure1}, these models produce nearly identical features regardless of temporal direction, confirming their representations are blind to the underlying procedural order. 
This demonstrates that while they can recognize an action (e.g., grinding beans), they fail to capture its essential temporal context (knowing it must come before brewing).
\begin{figure}[t!]
    \centering 

\begin{subfigure}{0.50\columnwidth}
        \centering
\includegraphics[width=\linewidth]{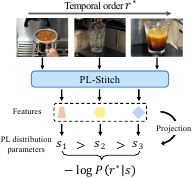}
        \caption{Learning via PL ranking}
        \label{fig:figure2a}
    \end{subfigure}
    \hfill \begin{subfigure}{0.49\columnwidth}
        \centering
        \includegraphics[width=\linewidth]{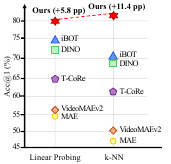}
        \caption{Results on phase recognition}
        \label{fig:figure2b}
    \end{subfigure}

\caption{\textbf{Core concept of the PL-Stitch model and key results.}
        (a) Our model, PL-Stitch, learns a procedurally-aware representation by optimizing for the negative log likelihood of the Plackett-Luce distribution, $-\log P(r^*|s)$, which maximizes the probability ($P$) of the ground-truth temporal order ($r^*$) given the model's predicted parameters ($s$).
(b) Significant performance gains are shown for the Cholec80 phase recognition task.}
    \label{fig:figure2_combined}
\end{figure}
This critical failure stems from the inherent design of prevalent SSL objectives, which are centered on local tasks like instance discrimination~\cite{Caron2021EmergingTransformers,Chen2020ARepresentations, Chen2020ImprovedLearning} or masked reconstruction~\cite{He2022MaskedLearners, HangboBaoandLiDongandSonghaoPiaoandFuruWei2022BEIT:TRANSFORMERS,Zhou2022IBOT:Tokenizer, Wang2023VideoMAEMasking}.
Consequently, despite achieving strong performance in static image analysis~\cite{Caron2021EmergingTransformers, Chen2020ImprovedLearning, Chen2020ARepresentations, He2022MaskedLearners, Zhou2022IBOT:Tokenizer} and short atomic clip modeling~\cite{Feichtenhofer2021ALearning, Wang2022BEVT:Transformers,ZhanTongandYibingSongandJueWangandLiminWang2022VideoMAE:Pre-Training, Wang2023VideoMAEMasking}, current SSL approaches remain fundamentally \textit{procedural-agnostic} (Fig.~\ref{fig:figure1a}).
They learn robust features for \textit{what} is in a frame, but not \textit{when} that frame occurs in the sequence.

To address this shortcoming, we propose \textbf{PL-Stitch}, a self-supervised framework that harnesses the inherent temporal order as a powerful supervisory signal (Fig.~\ref{fig:figure2a}). 
It is designed with two complementary branches.

The video branch serves as our main \emph{temporal ranking objective}. It learns global workflow progression by tasking the model to predict the correct chronological order of sampled frames.
Importantly, we formulate this as a listwise ranking problem using the probabilistic \textbf{Plackett-Luce (PL)} model~\cite{Plackett1975ThePermutations, Luce1977TheYears}. 
Our approach has \textbf{key advantages} over traditional temporal ordering tasks: 
1) Its listwise formulation is more efficient and globally consistent than suboptimal, pairwise comparisons~\cite{Hu2021ContrastLearning,Lee2017UnsupervisedSequences, Sermanet2017Time-ContrastiveObservation}, as it optimizes the ordering of a sequence of $k$ elements in a single step, providing a global signal rather than relying on $\mathcal{O}(k^2)$ fragmented local comparisons.
2) Its probabilistic, ranking-based nature is a more robust fit for modeling relative order than classification tasks~\cite{Misra2016ShuffleVerification, Wei2018LearningTime, Xu2019Self-SupervisedPrediction, Fernando2017Self-SupervisedNetworks}. 
Permutation classification tasks mistreat this relative problem by using absolute labels, penalizing near-correct orderings (minor sorting mistakes) as fully wrong.
In contrast, our PL-based ranking model computes a probability distribution over permutations, allowing penalties to scale with error severity.

The complementary image branch learns fine-grained, local features by jointly optimizing two objectives: 1) our novel spatio-temporal jigsaw objective, which uses adjacent past and future frames as temporal context to learn object correspondence, and 2) a masked image modeling (MIM) loss~\cite{Zhou2022IBOT:Tokenizer} for robust semantic representations.

Both video and image branch objectives are optimized jointly, compelling the encoder to learn a representation that is both semantically-rich and procedurally-aware.
As shown in Fig.~\ref{fig:figure2b}, PL-Stitch consistently outperforms state-of-the-art methods using the same backbone.

In summary, our \textbf{contributions} are as follows:
\begin{itemize}
    \item We identify and experimentally validate the procedural agnosticism of dominant SSL methods, demonstrating they are blind to the video's underlying procedural order.

\item To the best of our knowledge, we are the first to leverage the Plackett-Luce (PL) model to formulate probabilistic pretext tasks for self-supervised learning.

\item We propose two novel objectives based on the PL model within our PL-Stitch framework: a listwise temporal ranking objective to learn global workflow progression and a spatio-temporal jigsaw objective to capture fine-grained object correspondence. 

    \item We set a new state-of-the-art on five challenging surgical and cooking benchmarks, outperforming all baselines on both phase recognition and action segmentation tasks.

\end{itemize}

\begin{figure*}[t!]
	\centering
	\includegraphics[width=.9\linewidth]{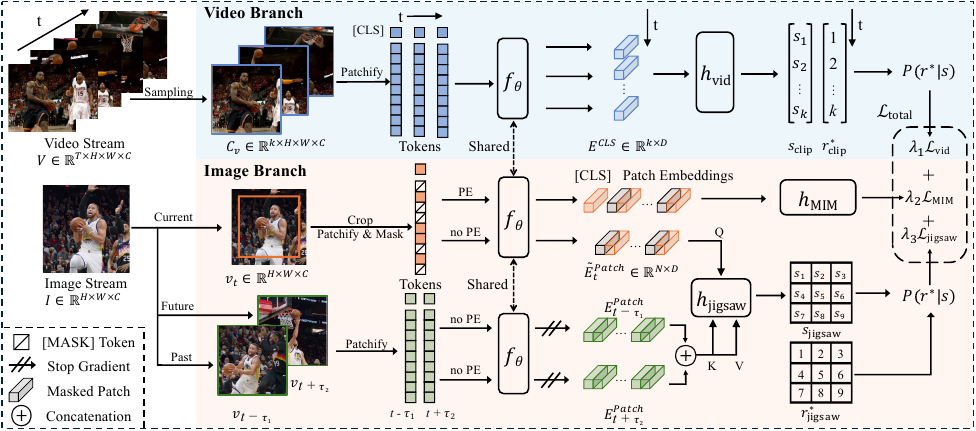}
    \caption{
        \textbf{Overview of the PL-Stitch framework.}
Our model jointly trains a shared backbone encoder ($f_{\theta}$) by using a \textbf{Video branch} (Sec.~\ref{sec:video_branch}) for global workflow progression and an \textbf{Image branch} (Sec.~\ref{sec:image_branch}) for fine-grained feature learning. 
The Video branch (top) treats time as order, training the encoder with a Plackett–Luce loss $\mathcal{L}_{\text{vid}}$ (Eq.~\ref{eq:pl_loss}, Eq.~\ref{eq:vid_loss}) to predict the correct relative chronological sequence of a sampled clip. 
The Image branch (bottom) learns robust local features by jointly optimizing a standard masked image modeling loss $\mathcal{L}_{\text{MIM}}$ with our novel spatio-temporal jigsaw $\mathcal{L}_{\text{jigsaw}}$, which learns object correspondence from adjacent frames (Eq.~\ref{eq:jigsaw_loss}).
The symbols $h_{\text{vid}}$, $h_{\text{MIM}}$, and $h_{\text{jigsaw}}$ denote task-specific projection heads.
By optimizing all objectives, the shared backbone learns a powerful representation sensitive to both procedural order and fine-grained visual details. \textbf{Best viewed online.}
}
	
	\label{fig:pipeline}
\end{figure*}

 \section{Related Work}
\label{sec:related_work}

\noindent 
\textbf{Self-supervised Visual Representation Learning.}
A widely adopted approach in self-supervised visual representation learning is contrastive learning, which seeks to learn an embedding space that groups similar images together.
Pioneer methods like MoCo~\cite{He2020MomentumLearning, Chen2020ImprovedLearning, Chen2021AnTransformers} and SimCLR~\cite{Chen2020ARepresentations} learn this space by pulling augmented \textit{positive} views of an image closer while pushing \textit{negative} views apart.
To prevent representational collapse without negative pairs, self-distillation methods like DINO~\cite{Caron2021EmergingTransformers} employ a student-teacher framework to match augmented views.
Furthermore, these contrastive and distillation approaches are also adapted to the video domain by leveraging temporal dynamics~\cite{Feichtenhofer2021ALearning, Recasens2021BroadenLearning, Qian2021SpatiotemporalLearning, Sun2023MaskedLearning, Qing2023Self-SupervisedConsistency, Wang2021EnhancingMotion}.
More recently, Masked Image Modeling (MIM) has become prominent, tasking a model with reconstructing masked patches at the pixel level (MAE~\cite{He2022MaskedLearners}) or latent-space level (BEiT~\cite{HangboBaoandLiDongandSonghaoPiaoandFuruWei2022BEIT:TRANSFORMERS}, iBOT~\cite{Zhou2022IBOT:Tokenizer}).
Building on the iBOT objective, DINOv2~\cite{Oquab2023DINOv2:Supervision} and DINOv3~\cite{Simeoni2025DINOv3} later scaled this approach with massive data to boost performance.
%
Masked reconstruction paradigms have also been successfully extended from images to higher-dimensional modalities~\cite{Wang2023VideoMAEMasking, ZhanTongandYibingSongandJueWangandLiminWang2022VideoMAE:Pre-Training, Pei2024VideoMAC:ConvNets, Sun2026Curve3D:Understanding, Wu2023DropMAE:Tasks, Feichtenhofer2022MaskedLearners, Wang2022BEVT:Transformers, Huang2023MGMAE:Autoencoding}, particularly videos.
Methods like VideoMAE~\cite{ZhanTongandYibingSongandJueWangandLiminWang2022VideoMAE:Pre-Training} learn representations by reconstructing randomly masked spatio-temporal tubes from video sequences.
However, a key limitation is its symmetric treatment of space and time~\cite{Feichtenhofer2022MaskedLearners,Huang2023MGMAE:Autoencoding}, which ignores the inherent causal progression of the temporal dimension and makes such methods suboptimal for learning the dynamics of procedural events.

\noindent \textbf{Temporal Correspondence Learning.}
Understanding temporal correspondence between frames is crucial for modeling procedural activities.
Early supervised methods using optical flow and motion estimation were effective but required costly pixel-level annotations~\cite{Teed2020RAFT:Flow, Sun2018PWC-Net:Volume, Xu2017AccurateProcessing, Ilg2017FlowNetNetworks}.
To overcome this issue, self-supervised approaches learned spatio-temporal correspondence by tracking objects or patches across frames~\cite{Wang2019LearningTime, Zhang2023BoostingLearning, Jabri2020Space-timeWalk, Xu2021RethinkingPerspective, Li2019Joint-taskCorrespondence}. 
More recently, temporal predictive reconstruction has emerged, where SiamMAE~\cite{Gupta2023SiameseAutoencoders} predicts a future frame from the past and T-CoRe~\cite{Liu2025WhenLearning} using both past and future frames to reconstruct a central frame.
While these methods excel at learning fine-grained correspondence, they lack a mechanism to capture video's long-term procedural structure.
Training models to order frame sequences has also been proposed~\cite{Xu2019Self-SupervisedPrediction, Wei2018LearningTime, Lee2017UnsupervisedSequences, Fernando2017Self-SupervisedNetworks, Misra2016ShuffleVerification, Hu2021ContrastLearning, Sermanet2017Time-ContrastiveObservation}. 
However, this approach is limited by suboptimal objectives, such as pairwise comparisons that provide a local signal~\cite{Hu2021ContrastLearning,Lee2017UnsupervisedSequences, Sermanet2017Time-ContrastiveObservation}, or permutation classification tasks that miscast relative ordering as absolute classification~\cite{Xu2019Self-SupervisedPrediction, Wei2018LearningTime, Misra2016ShuffleVerification}.
Our work, PL-Stitch, addresses this gap by reformulating temporal understanding as a more robust, probabilistic listwise ranking problem, which we argue is a more natural and direct way to model sequences.

\begin{figure}[!t]
	\centering
	\includegraphics[width=.87\linewidth]{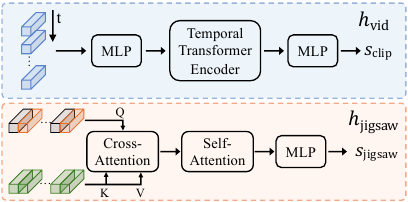}
    \caption{
        \textbf{Structure of our $h_{\text{vid}}$ and $h_{\text{jigsaw}}$ heads.}
The video head ($h_{\text{vid}}$) consists of an MLP to reduce feature dimensionality for computational efficiency, a Transformer Encoder to aggregate global context across the $k$ frame features for ordering, and a final MLP that outputs the PL distribution parameters $s_{\text{clip}}$.
The jigsaw head ($h_{\text{jigsaw}}$) uses Cross-Attention to aggregate temporal context (K, V) onto the target patches (Q), followed by Self-Attention for spatial relationships refinement, and a final MLP for producing PL distribution parameters $s_{\text{jigsaw}}$.  
        }
	\label{fig:heads}
\end{figure}


\noindent
\textbf{Procedural Activity Understanding.}
Modeling procedural activities, which involves both recognizing action steps and their temporal sequence, is a long-standing challenge in video understanding. Most recent works consider training on densely-annotated videos to localize and predict the order of steps and the logical dependencies~\cite{Xie2026SpatiaLQA:Models, Chang2020ProcedureVideos, Kuehne2014TheActivities, Tang2019COIN:Analysis, Zhukov2019Cross-TaskVideos, Damen2021TheBaselines}. To reduce this reliance on expensive labels, other methods have explored weakly-supervised approaches, such as learning from only a transcript of action steps without timestamps~\cite{Bojanowski2015Weakly-SupervisedText, Zhao2022PSupervision,Zhukov2019Cross-TaskVideos}.
In contrast, our work operates in a fully self-supervised regime, learning a representation from the video signal alone that can be leveraged for procedural video understanding.
\section{Methodology}
\label{sec:method}

\subsection{Problem Formulation}
\label{sec:problem_formulation}
Our primary objective is to pretrain a frame-level feature encoder, $f_\theta$, without relying on manual annotations. 
An input video is a sequence $V = (v_1, v_2, \dots, v_T)$ of $T$ frames, where each frame $v_t \in \mathbb{R}^{H \times W \times C}$. 
Following the standard Vision Transformer (ViT) paradigm, the encoder first tokenizes a frame $v_t$ by splitting it into a grid of non-overlapping patches, each of size $P \times P$.
These patches are first projected into $D$-dimensional patch tokens.
A learnable \texttt{[CLS]} token is prepended to this sequence to form the initial input tokens, which we denote as $Z \in \mathbb{R}^{(N+1) \times D}$, where $N=H \times W/P^2$.
The token sequence $Z$ is then processed by the Transformer blocks to produce the final, context-aware output embeddings, denoted as $E \in \mathbb{R}^{(N+1) \times D}$. 
This completes the full mapping $f_\theta: \mathbb{R}^{H \times W \times C} \to \mathbb{R}^{(N+1) \times D}$.

\begin{table*}[!t]
    \centering
    \small
    \setlength{\tabcolsep}{1.9pt}
    \caption{
        \textbf{Linear probing and \textit{k}-NN classification results for surgical phase recognition task.} 
        We report top-1 accuracy and F1-score for linear probing and top-1 accuracy for \textit{k}-NN evaluations (\textit{k} = 20) on the AutoLaparo, Cholec80, and M2CAI16 dataset.
The experiments are conducted with frozen backbone to demonstrate the method's effectiveness. The predictions are computed on a frame-by-frame basis for all tasks.
Type `S' denotes a surgical-specific foundation model and `G' denotes a generalist self-supervised method.
Best in \textbf{bold}.
        }
    \label{tab:linear_probing_classification}
    \begin{tabularx}{\linewidth}{@{}lcccccccccccccccccc@{}}
    \hline
        \multirow{3}{*}{Method}  &\multirow{3}{*}{Type}  &\multirow{3}{*}{Pretraining data} &\multirow{3}{*}{Backbone} &~ & \multicolumn{4}{c}{AutoLaparo} & ~ & \multicolumn{4}{c}{Cholec80}  & ~ & \multicolumn{4}{c}{M2CAI16}\\
        \cline{6-9} 
        \cline{11-14} 
        \cline{16-19} 
        & ~ & ~& ~ &~ &  \multicolumn{2}{c}{Linear} &~ &\textit{k}-NN
        & ~  &  \multicolumn{2}{c}{Linear} & ~&\textit{k}-NN
        & ~  &  \multicolumn{2}{c}{Linear} & ~&\textit{k}-NN\\
        \cline{6-7} 
        \cline{9-9} 
        \cline{11-12}
        \cline{14-14}
        \cline{16-17}
        \cline{19-19}
        & ~ & ~ & ~ & ~ &  Acc &  F1-score & ~ & Acc
        & ~  &  Acc &  F1-score & ~ & Acc
        & ~  &  Acc &  F1-score & ~ & Acc\\
        \hline

        Endo-FM~\cite{Wang2023FoundationPre-train} &\multirow{3}{*}{S} & Private data  & ViT-B/16 & ~
        & $51.5$ & $43.1$ & ~ & $46.8$
        & ~ 
        & $62.7$  & $53.9$ & ~ & $55.9$
        & ~ 
        & $53.8$ & $49.8$ & ~ & $44.8$
    
        \\ 
        EndoViT~\cite{Batic2024EndoViT:Images} & ~ & Merged public data  & ViT-B/16 & ~
        & $45.4$ & $38.7$ & ~ & $40.2$
        & ~ 
        & $46.9$  & $39.0$ & ~ & $45.8$
        & ~ 
        & $38.5$ & $37.2$ & ~ & $34.6$
        \\ 

        LemonFM~\cite{Che2025LEMON:Settings}  & ~ & LEMON  & ConvNeXt-B & ~

        & $74.7$ & $64.5$ & ~ & $73.1$
        & ~ 
        & $73.9$  & $65.8$ & ~ & $69.5$
        & ~ 
        & $68.4$ & $62.4$ & ~ & $65.8$
        \\
        \hline

        MAE~\cite{He2022MaskedLearners} &\multirow{6}{*}{G} & LEMON  & ViT-B/16 &  ~
        & $35.5$ & $32.0$ & ~ & $35.9$
        & ~ 
        & $54.9$  & $43.4$ & ~ & $47.2$
        & ~ 
        & $40.7$ & $38.1$ & ~ & $36.4$
         \\  

        VideoMAEv2~\cite{Wang2023VideoMAEMasking} & ~ & LEMON  & ViT-B/16 & ~
        & $49.8$ & $42.4$ & ~ & $46.8$
        & ~ 
        & $55.8$  & $48.5$ & ~ & $50.0$
        & ~ 
        & $50.7$ & $41.3$ & ~ & $45.8$
         \\  

        DINO~\cite{Caron2021EmergingTransformers} & ~ & LEMON  & ViT-B/16 & ~

        & $74.9$ & $65.0$ & ~ & $72.8$
        & ~ 
        & $72.2$  & $67.1$ & ~ & $69.4$
        & ~ 
        & $67.6$ & $62.8$ & ~ & $65.2$
        \\

        iBOT~\cite{Zhou2022IBOT:Tokenizer}  & ~ & LEMON  & ViT-B/16 & ~

        & $76.3$ & $65.1$ & ~ & $75.3$
        & ~ 
        & $74.6$  & $67.6$ & ~ & $70.3$
        & ~ 
        & $71.0$ & $64.5$ & ~ & $68.0$
        \\
        
        T-CoRe~\cite{Liu2025WhenLearning} & ~ & LEMON  & ViT-B/16 & ~

        & $67.9$ & $58.0$ & ~ & $61.8$
        & ~ 
        & $65.1$  & $59.4$ & ~ & $62.3$
        & ~ 
        & $61.8$ & $57.1$ & ~ & $60.2$
        \\

        \textbf{PL-Stitch (ours)} & ~ & LEMON  & ViT-B/16 & ~
        & $\mathbf{79.9}$ & $\mathbf{69.0}$ & ~ & $\mathbf{82.5}$
        & ~ 
        & $\mathbf{80.4}$  & $\mathbf{73.0}$ & ~ & $\mathbf{81.7}$
        & ~ 
        & $\mathbf{76.4}$ & $\mathbf{69.1}$ & ~ & $\mathbf{77.1}$
        \\
        \hline

    \end{tabularx}
\end{table*}

\subsection{Overview}
\label{sec:overview}
Our goal is to learn a self-supervised representation capable of capturing both procedural temporal correspondence and fine-grained spatio-temporal relationships. To that end, we propose PL-Stitch, a multi-task framework that jointly trains a shared backbone encoder, $f_\theta$, using two complementary branches as illustrated in Fig.~\ref{fig:pipeline}. 
The \textbf{Video branch} (Sec.~\ref{sec:video_branch}) is designed to learn temporal ordering across variable time scales by training the encoder to predict the correct chronological sequence of a sampled clip ($\mathcal{L}_{\text{vid}}$).
In parallel, the \textbf{Image branch} (Sec.~\ref{sec:image_branch}) learns robust local features through two tasks: our novel spatio-temporal jigsaw ($\mathcal{L}_{\text{jigsaw}}$), which learns fine-grained correspondence using adjacent frames as context and a masked image modeling objective ($\mathcal{L}_{\text{MIM}}$) based on iBOT~\cite{Zhou2022IBOT:Tokenizer}.
Importantly, both the \textbf{temporal ordering} and the \textbf{spatio-temporal jigsaw} tasks are framed as listwise ranking problems, optimized with a unified \textbf{Plackett-Luce} objective (Sec.~\ref{sec:pl_formulation}). 
The entire training procedure is summarized in Algorithm~\ref{alg:PL-Stitch}.

\begin{algorithm}[t!]
    \caption{PL-Stitch Optimization Algorithm}
    \label{alg:PL-Stitch}
    \begin{algorithmic}[1]
\Statex \textbf{Input:} Unlabeled dataset $\mathcal{D}$.
        \Statex \textbf{Parameters:} Batch size $B$, loss weights $\lambda_1, \lambda_2, \lambda_3$, and maximum iterations $L = |\mathcal{D}| / B$.
        \Statex \textbf{Initialize:} Encoder $f_\theta$ and heads $h_{\text{vid}}, h_{\text{MIM}}, h_{\text{jigsaw}}$.
        
        \Statex
\Procedure{PL-Stitch\_Step}{$V_m, I_m$}
            \Statex {\textit{\textcolor{videoblue}{Video Branch}}}
\State Sample sparse clip $C_v = (v_{1}, \dots, v_{k})$ from $V_m$.
            \State Define target order $r^*_{\text{clip}} = (1, 2, \dots, k)$.
            \State Predict $s_{\text{clip}} = h_{\text{vid}}(f_\theta(C_v))$.
            \State Compute $\mathcal{L}_{\text{vid}} = - \log P(r^*_{\text{clip}} | s_{\text{clip}})$ by Eq.~\ref{eq:vid_loss}.
            
            \Statex {\textit{\textcolor{imageorange}{Image Branch}}}
\State Sample frame triplet $(v_{t-\tau_1}, v_t, v_{t+\tau_2})$ from $I_m$ and mask $v_t \to \tilde{v}_t$.
            \State Compute MIM loss $\mathcal{L}_{\text{MIM}}$ from $\tilde{v}_t, v_t$.
            \State Compute jigsaw query Q from $\tilde{v}_t$; context K, V from $v_{t-\tau_1}, v_{t+\tau_2}$ (no PE).
            \State Define target order $r^*_{\text{jigsaw}} = (1, 2, \dots, N)$.
            \State Predict $s_{\text{jigsaw}} = h_{\text{jigsaw}}(Q, K, V)$.
            \State Compute $\mathcal{L}_{\text{jigsaw}} = - \log P(r^*_{\text{jigsaw}} | s_{\text{jigsaw}})$ by Eq.~\ref{eq:jigsaw_loss}.

            \State $\mathcal{L}_{\text{total}} = \lambda_1 \mathcal{L}_{\text{vid}} + \lambda_2 \mathcal{L}_{\text{MIM}} + \lambda_3 \mathcal{L}_{\text{jigsaw}}$ by Eq.~\ref{eq:total_loss}.
            \State Update $\theta$ using $\nabla_{\theta}\mathcal{L}_{\text{total}}$.
        \EndProcedure
        
        \Statex
        \For{$l=1$ \textbf{to} $L$}
            \State Sample a batch of data $\{(V_i, I_i)\}_{i=1}^B$ from $\mathcal{D}$.
            \For{$m=1$ \textbf{to} $B$}
                \State \Call{PL-Stitch\_Step}{$V_m, I_m$} \Comment{\textcolor{blue!80}{Parallelizable}}
            \EndFor
        \EndFor
        \State \textbf{return} Pre-trained encoder parameters $\theta$.
\end{algorithmic}
\end{algorithm}

\subsection{Plackett-Luce Ranking Formulation}
\label{sec:pl_formulation}
Both our temporal and spatial learning tasks are framed as a listwise ranking problem. 
We address this using the Plackett-Luce (PL) distribution~\cite{Plackett1975ThePermutations, Luce1977TheYears}, a probabilistic framework for defining a distribution over all $K!$ possible permutations of $K$ items.
The PL model is derived from Luce’s Choice Axiom~\cite{Luce1977TheYears} (independence from irrelevant alternatives), which states that the odds of choosing item $i$ over item $j$ depend only on those two items. 
This assumption aligns well with relative time ordering, where the preference between two frames is invariant to the rest of the sequence.
Given a set of $K$ items, the PL model is parameterized by a vector of real-valued scores $s \in \mathbb{R}^K$. 
The probability of observing a specific permutation $r$ of the set $\{1, \dots, K\}$ is given by the likelihood:
\begin{align}
    P(r | s) = \prod_{i=1}^{K} \frac{\exp(s_{r(i)})}{\sum_{j=i}^{K} \exp(s_{r(j)})}
    \label{eq:pl_prob}
\end{align}
where $r(i)$ denotes the value of the item placed at position $i$ in the permutation $r$.
In our work, we train our self-supervised model to predict the vector of parameters $s$ for a given set of $K$ items (i.e., video frames for temporal ordering or image patches for jigsaw ranking).
The network is trained via Maximum Likelihood Estimation (MLE) to learn parameters $s$ that maximize the probability of the single correct ground-truth permutation $r^{*}$.
This is achieved by minimizing the negative log-likelihood, which serves as our general ranking loss, $\mathcal{L_{\text{PL}}}$:
\begin{align}
    \label{eq:pl_loss}
    \mathcal{L}_{\text{PL}}(s, r^*) = - \log P(r^* | s).
\end{align}
This formulation is applied to both the video and jigsaw branches, providing a consistent optimization target for learning ordered sequences.

\subsection{Video Branch: Listwise Temporal Ranking}
\label{sec:video_branch}
The Video branch (top of Fig.~\ref{fig:pipeline}) learns a representation of procedural progression through a PL-based listwise ranking loss that trains the encoder to correctly order a sequence of frames.
From a video $V$ of length $T$, we sample a clip $C_v = (v_{t_0}, v_{t_0+\Delta t}, \dots, v_{t_0+(k - 1)\Delta t})$ containing $k$ frames, with a random start time $t_0$ and step size $\Delta t \ge 1$ such that $t_0 + (k - 1)\Delta t \le T$. The ground-truth chronological order for this clip is represented by \(r^*_{\text{clip}} = (1, 2, \dots, k)\).

Each of the $k$ frames is independently processed by the shared encoder $f_\theta$, and for computational efficiency, we feed only their \texttt{[CLS]} embeddings ($E^{\text{CLS}}\in\mathbb{R}^{k \times D}$) into the temporal head $h_{\text{vid}}$, avoiding the costly fusion of numerous patch embeddings.
As illustrated in Fig.~\ref{fig:heads}, this head
outputs a vector of estimated PL distribution parameters $s_{\text{clip}} = (s_1, \dots, s_k)$.
These parameters are then trained using the Plackett-Luce objective (Sec.~\ref{sec:pl_formulation}), making the video branch loss an instantiation of Eq.~\ref{eq:pl_loss}:
\begin{equation}
    \mathcal{L}_{\text{vid}} = \mathcal{L}_{\text{PL}}(s_{\text{clip}}, r^*_{\text{clip}}).
    \label{eq:vid_loss}
\end{equation}

\begin{table}[t!]
    \centering
    \setlength{\tabcolsep}{5pt} 
    
    \caption{
        \textbf{Comparison with state-of-the-art self-supervised methods on cooking datasets.}
We use the egocentric cooking dataset GTEA~\cite{Fathi2011LearningActivities} and the third-person cooking dataset Breakfast~\cite{Kuehne2014TheActivities}.
All methods are evaluated on a ViT-B/16 backbone using the linear probing and $k$-NN protocols.
    }
    \label{tab:food_exp}
    \renewcommand{\arraystretch}{1.05}
    
\resizebox{\columnwidth}{!}{

    \begin{tabular}{lcc c c c} 
        \toprule
        \multirow{2}{*}{Method} & \multicolumn{3}{c}{Linear Probing} & & $k$-NN \\
        \cline{2-4} \cline{6-6}
        & Acc & Edit & F1@\{10,25,50\} & & Acc \\
        \midrule

\rowcolor{MyBlue!70} 
        \multicolumn{6}{l}{\textbf{GTEA}} \\
        VideoMAEv2~\cite{Wang2023VideoMAEMasking}    & 
        $30.2$   & $22.8$ & $23.5$\; $17.8$\; $10.0$ & & $26.3$ \\
        
        DINO~\cite{Caron2021EmergingTransformers} & 
        $52.2$   & $53.9$ & $55.3$\; $46.7$\; $28.3$ & & $60.0$ \\
        
        iBOT~\cite{Zhou2022IBOT:Tokenizer}   & 
        $51.3$ & $57.2$ & $59.3$\; $49.0$\; $27.4$ & & $59.2$ \\
        
        T-CoRe~\cite{Liu2025WhenLearning}       & 
        $47.2$ & $50.0$ & $49.3$\; $42.8$\; $20.9$ & & $42.7$ \\
        
        \textbf{PL-Stitch (ours)}   & 
        $\mathbf{54.1}$ & $\mathbf{60.2}$ & $\mathbf{61.7}$\; $\mathbf{50.0}$\; $\mathbf{30.4}$ & & $\mathbf{62.4}$ \\
        \midrule

\rowcolor{MyBlue!70} 
        \multicolumn{6}{l}{\textbf{Breakfast}} \\
        VideoMAEv2~\cite{Wang2023VideoMAEMasking}    & 
        $12.3$   & $14.7$ & $8.3$\; $4.9$\; $2.2$ & & $4.6$ \\
        
        DINO~\cite{Caron2021EmergingTransformers}  & 
        $15.9$   & $13.1$ & $8.2$\; $5.6$\; $\mathbf{3.6}$ & & $7.1$ \\
        
        iBOT~\cite{Zhou2022IBOT:Tokenizer}  & 
        $15.7$ & $12.0$ & $7.2$\; $4.8$\; $2.6$ & & $7.5$ \\
        
        T-CoRe~\cite{Liu2025WhenLearning}  & 
        $12.0$ & $12.5$ & $6.1$\; $4.0$\; $1.8$ & & $6.1$ \\
        
        \textbf{PL-Stitch (ours)}  & 
        $\mathbf{21.6}$ & $\mathbf{15.1}$ & $\mathbf{9.0}$\; $\mathbf{6.3}$\; $3.2$ & & $\mathbf{10.9}$ \\
        \bottomrule
    \end{tabular}
    
} \end{table}

\subsection{Image Branch: Local and Spatio-temporal Learning}
\label{sec:image_branch}
The Image branch (bottom of Fig.~\ref{fig:pipeline}) learns fine-grained features from local temporal context by operating on a triplet of frames, $(v_{t - \tau_1}, v_t, v_{t + \tau_2})$
, where $\tau_1,\tau_2$ are short temporal offsets.
The branch comprises two parallel objectives.

\noindent
\textbf{Masked Image Modeling.} We adopt the masked image modeling (MIM) objective from iBOT~\cite{Zhou2022IBOT:Tokenizer} on the current frame $v_t$ to establish a robust frame-level feature representation, optimized with the loss $\mathcal{L}_{\text{MIM}}$.

\noindent 
\textbf{Spatio-temporal Jigsaw.}
To learn fine-grained object correspondence, we introduce a jigsaw task guided by temporal context from neighboring frames.
The model must infer the original spatial arrangement of a central frame, which is tokenized into a sequence of $N$ patch tokens, some of which are masked.
The task is handled by our jigsaw head, $h_{\text{jigsaw}}$, which processes three sets of patch embeddings obtained from the shared backbone encoder, $f_\theta$.
The patch embeddings from the masked current frame, $\tilde{E}_t^{\text{patch}} \in \mathbb{R}^{N \times D}$, serve as the \textbf{Queries (Q)}, while the concatenated embeddings from the past and future frames, $[E_{t-\tau_1}^{\text{patch}}; E_{t+\tau_2}^{\text{patch}}]$, serve as the \textbf{Keys (K)} and \textbf{Values (V)}. 
To ensure reliance on visual content, positional embeddings are omitted.
As shown in Fig.~\ref{fig:heads}, the jigsaw head
produces a vector of estimated PL parameters $s_{\text{jigsaw}} = (s_1, \dots, s_N)$. 
We train the model to predict the original linearized patch order, $r^*_{\text{jigsaw}} = (1, 2, \dots, N)$, by minimizing the jigsaw loss $\mathcal{L}_{\text{jigsaw}}$ using the Plackett-Luce objective defined in Eq.~\ref{eq:pl_loss}.
\begin{equation}
    \mathcal{L}_{\text{jigsaw}} = \mathcal{L}_{\text{PL}}(s_{\text{jigsaw}}, r^*_{\text{jigsaw}}).
    \label{eq:jigsaw_loss}
\end{equation}

\subsection{Total Objective}
\label{sec:total_loss}

The final training objective for our PL-Stitch framework is the weighted sum of the three losses:
\begin{equation}
    \mathcal{L}_{\text{total}} = \lambda_1 \mathcal{L}_{\text{vid}} + \lambda_2 \mathcal{L}_{\text{MIM}} + \lambda_3 \mathcal{L}_{\text{jigsaw}} 
    \label{eq:total_loss}
\end{equation}
where $\lambda_1$, $\lambda_2$, and $\lambda_3$ are hyperparameters balancing each task’s contribution.
This final objective is used to update the network parameters during optimization (Algorithm~\ref{alg:PL-Stitch}).

 \section{Experiments}
\label{sec:evaluation}
\begin{table}[t!]
    \centering
    \small
    \setlength{\tabcolsep}{4.8pt}
    \caption{
        \textbf{Ablation on components of PL-Stitch.}
        We report top-1 accuracy on linear probing and \textit{k}-NN results for Cholec80 phase recognition.
    }
    \label{tab:ablation_component}
    \begin{tabularx}{.7\linewidth}{@{}cccccc@{}}
    \hline
        \multirow{1}{*}{No.} & \multirow{1}{*}{$\mathcal{L}_{\text{MIM}}$}  & \multirow{1}{*}{$\mathcal{L}_{\text{vid}}$}  & \multirow{1}{*}{$\mathcal{L}_{\text{jigsaw}}$} & Linear & \textit{k}-NN
        \\
        \cline{5-6} 

    \hline 
    
     1
        & \checkmark
        & 
        &
        & $73.4$  
        &  $69.4$   \\
    
    2
        & \checkmark
        & \checkmark
        &  
        &  $77.1$  
        &  $78.9$  
    \\
    3
        & \checkmark
        & 
        & \checkmark 
        &  $75.3$  
        &  $71.4$  
    \\

4
        & \checkmark \cellcolor{Gray!20}
        & \checkmark \cellcolor{Gray!20}
        & \checkmark \cellcolor{Gray!20}
        &  $\mathbf{77.8}$  \cellcolor{Gray!20}
        &  $\mathbf{80.2}$  \cellcolor{Gray!20}
    \\

    \hline

    \end{tabularx}
\end{table}

\begin{table}[t!]
    \centering
    \small
    \setlength{\tabcolsep}{4pt} \caption{
        \textbf{Ablation on temporal objective formulation.} We replace our $\mathcal{L}_{\text{vid}}$ (PL) with simpler objectives (row 2 from Table~\ref{tab:ablation_component}).
    }
    \label{tab:ablation_loss_type}
    \begin{tabularx}{.7\linewidth}{@{}cccc@{}}
    \hline
        No. & Temporal Objective & Linear & \textit{k}-NN \\
    \hline
        1 & Pairwise~\cite{Hu2021ContrastLearning} & $75.8$ & $75.4$ \\
        2 & Permutation (CE)~\cite{Misra2016ShuffleVerification} & $74.5$ & $70.1$ \\
        3 & PL Ranking (Eq.~\ref{eq:vid_loss}) \cellcolor{Gray!20} & $\mathbf{77.1}$\cellcolor{Gray!20} & $\mathbf{78.9}$\cellcolor{Gray!20} \\
    \hline
    \end{tabularx}
\end{table}

\begin{table}[t!]
    \centering
    \small
    \setlength{\tabcolsep}{10pt} \caption{
        \textbf{Ablation on number of temporal frames ($k$)} for $\mathcal{L}_{\text{vid}}$.
    }
    \label{tab:ablation_frames}
\begin{tabularx}{.7\linewidth}{@{}cccc@{}} \hline
No. & Frames ($k$) & Linear & \textit{k}-NN \\
    \hline
1 & $4$ & $76.5$ & $78.1$ \\
        2  & $8$\cellcolor{Gray!20} & $\mathbf{77.8}$\cellcolor{Gray!20} & $80.2$\cellcolor{Gray!20} \\
        3 & $12$ & $77.2$ & $79.5$ \\
        4 & $16$ & $77.5$ & $\mathbf{80.3}$ \\
    \hline
    \end{tabularx}
\end{table}
In this section, we first detail the experimental setup in Sec.~\ref{sec:exp_setup}.
Next, we provide quantitative comparisons across five datasets in Sec.~\ref{sec:sota_compare}.
Finally, we present ablation studies in Sec.~\ref{sec:ablation} and qualitative results in Sec.~\ref{sec:qualitative} (see the supplementary material for more details).

\subsection{Experimental Setup}
\label{sec:exp_setup}
\noindent 
\textbf{Datasets and Evaluation Protocols.} 
We evaluate on five widely used datasets from the surgical and cooking domains, selected for their long-range, multi-step procedural workflows with fixed sequential dependencies. 
For surgical procedures, we evaluate the temporal phase recognition task using AutoLaparo~\cite{Wang2022AutoLaparo:Hysterectomy} (laparoscopic hysterectomy), along with Cholec80~\cite{Twinanda2017EndoNet:Videos} and M2CAI16~\cite{Stauder2016TheChallenge} (both laparoscopic cholecystectomy).
For cooking activities, we evaluate the temporal action segmentation task using GTEA~\cite{Fathi2011LearningActivities} (egocentric cooking) and Breakfast~\cite{Kuehne2014TheActivities} (third-person cooking).
We assess feature quality using two standard protocols: \textbf{Linear Probing}, which trains a linear classifier on frozen features, and \textbf{\textit{k}-NN Classification}~\cite{Caron2021EmergingTransformers}, which uses $k=20$ nearest neighbors in the feature space for non-parametric evaluation.
For \textit{k}-NN, we report top-1 accuracy. 
For Linear probing, we follow established benchmarks to report frame-wise accuracy and F1-score for surgical datasets~\cite{Che2025LEMON:Settings, Wang2022AutoLaparo:Hysterectomy, Liu2025LoViT:Recognition}, and frame-wise accuracy, segmental edit distance, and the segmental F1 score at overlapping thresholds 10\%, 25\% and 50\% for cooking datasets~\cite{Li2023MS-TCN++:Segmentation, Fathi2011LearningActivities, Kuehne2014TheActivities}.

\begin{figure}[t!]
	\centering
	\includegraphics[width=.97\linewidth]{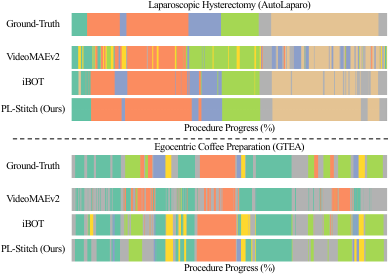}
	\caption{
        \textbf{Visualization of linear probing} predictions for phase recognition (top) and action segmentation (bottom). Each horizontal bar shows the frame-wise predictions, where the x-axis denotes progress over time and the colors represent different classes.
}
	\label{fig:progress_qualitative}
\end{figure}

\begin{figure*}[t!]
	\centering
	\includegraphics[width=.86\linewidth]{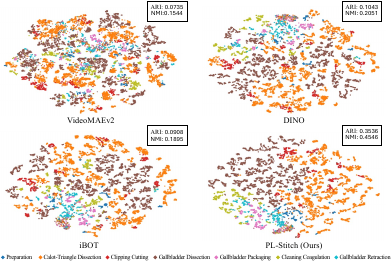}
	\caption{
    \textbf{t-SNE visualization} of frozen backbone features for Cholec80 phase recognition. The plot also reports the clustering quality metrics Adjusted Rand Index (ARI) and Normalized Mutual Information (NMI), where higher values are better. Our PL-Stitch model demonstrates superior class separation (ARI: 0.3536, NMI: 0.4546) compared to VideoMAEv2, DINO, and iBOT.
    }
	\label{fig:tsne}
\end{figure*}

\noindent
\textbf{Pretraining Details.}
We adopt ViT-B/16~\cite{Dosovitskiy2020AnScale} as our framework's backbone.
For our temporal ranking objective ($\mathcal{L}_{\text{vid}}$), we sample \textbf{k=8} frames from each video of length $T$. 
We first select a random starting frame $t_0$ and then a random, uniform step length $\Delta t$, such that $t_0 + 7\Delta t \le T$.
Videos are resampled multiple times per epoch, providing broad coverage of the procedural structure.
For the $\mathcal{L}_{\text{jigsaw}}$ objective, the past and future context frames are sampled from the temporal ranges [-2.5, -1.5]s and [+1.5, +2.5]s relative to the current frame, following~\cite{Liu2025WhenLearning}.
For the $\mathcal{L}_{\text{MIM}}$ objective, we apply block-wise masking with a ratio of 30\% as in~\cite{Zhou2022IBOT:Tokenizer}.
The loss weights (Eq.~\ref{eq:total_loss}) are set to $\lambda_1=1$, $\lambda_2=1$, and $\lambda_3=0.4$ with analysis in supplementary material.

\noindent
\textbf{Optimization Strategies.} 
We employ different pretraining optimization strategies based on the target domain.
For surgical datasets, models are pretrained for 30 epochs on the large-scale surgical video dataset LEMON~\cite{Che2025LEMON:Settings} to leverage its domain-specific knowledge. 
We use AdamW with a base learning rate of $4 \times 10^{-4}$, a 3-epoch cosine warmup, and a total batch size of 240. 
For cooking datasets, we pretrain directly on their respective official training sets for 100 epochs with 10 warmup epochs, as no comparable large-scale pretraining dataset exists for this highly variable domain; other optimizer settings are kept identical.
Experiments utilize 4 NVIDIA A100 (40GB) GPUs and PyTorch 2.5.1~\cite{Paszke2019PyTorch:Library}.

\begin{figure*}[t!]
	\centering
	\includegraphics[width=.81\linewidth]{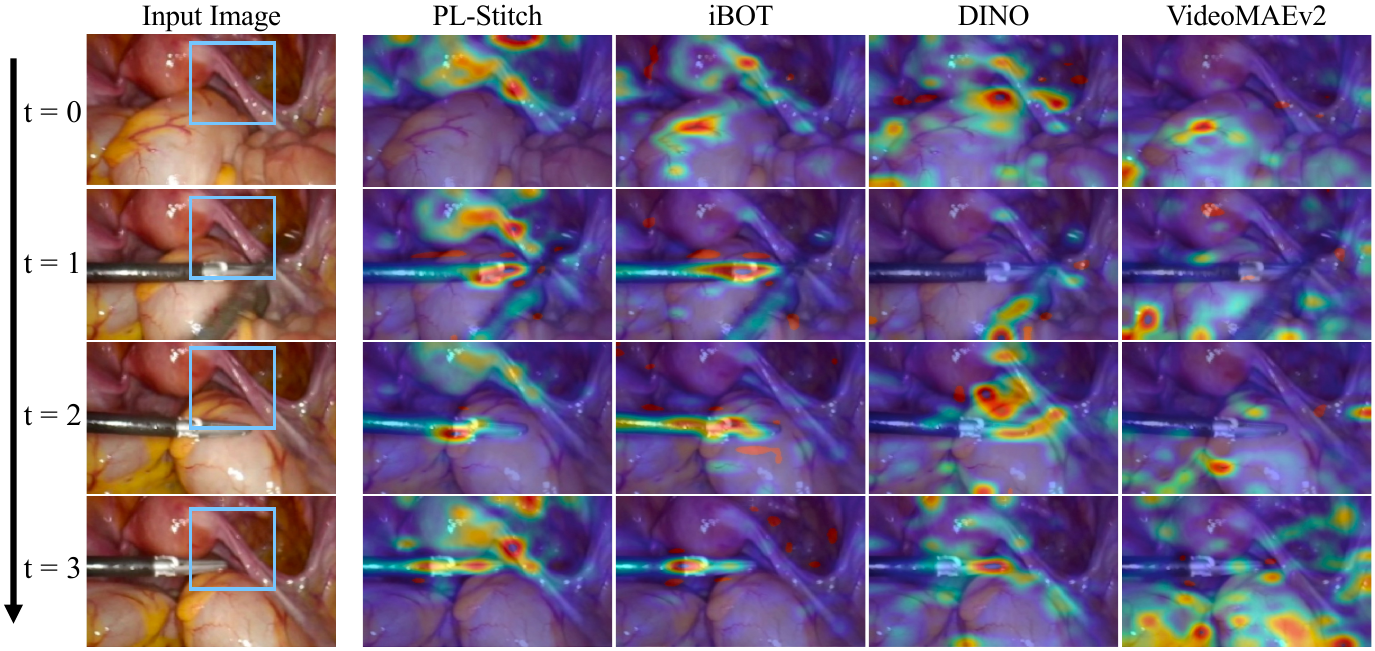}
	\caption{
    \textbf{Attention maps} queried by the [CLS] token comparing our method, PL-Stitch, against DINO, iBOT, and VideoMAEv2 on a sequence of frames from AutoLaparo~\cite{Wang2022AutoLaparo:Hysterectomy} during the \textit{Dividing Ligament and Peritoneum} phase. The \textcolor{cyan}{light blue} box indicates the instrument operating area, which is critical for understanding the surgical phase. While other methods produce diffuse attention that changes erratically over time, our PL-Stitch maintains a consistent and accurate focus on the target area (the instrument and its operating area).
}

	\label{fig:attention}
\end{figure*}

\subsection{Main Quantitative Results}
\label{sec:sota_compare}
We provide a comprehensive comparison of PL-Stitch against leading self-supervised models.

\noindent
\textbf{Surgical Phase Recognition.}
As shown in Table~\ref{tab:linear_probing_classification}, PL-Stitch consistently outperforms all baselines on AutoLaparo, Cholec80, and M2CAI16 under both linear probing and \textit{k}-NN evaluation.
This demonstrates superior feature quality over both generalist (G) and specialist (S) methods. The performance gains are particularly striking in the \textit{k}-NN evaluation, which directly assesses feature space quality. 
On Cholec80, PL-Stitch achieves 81.7\% k-NN accuracy, a significant \textbf{+11.4 pp} gain over the strong iBOT baseline~\cite{Zhou2022IBOT:Tokenizer}. This trend holds for AutoLaparo (+7.2 pp) and M2CAI16 (+9.1 pp) as well. Furthermore, PL-Stitch also achieves the highest scores in all linear probing metrics, underscoring the discriminative power of its frozen features.

\noindent
\textbf{Cooking Action Segmentation.}
We validate our model’s generalizability on the cooking datasets in Table~\ref{tab:food_exp}.
On GTEA, PL-Stitch surpasses the best baselines in \textit{k}-NN accuracy (62.4\% vs. 60.0\%) and in linear probing, with higher accuracy (54.1\% vs. 52.2\%), Edit score (60.2\% vs. 57.2\%), and F1@10 (61.7\% vs. 59.3\%).
On Breakfast, this superiority is even more pronounced with a linear probing accuracy gain of \textbf{+5.7 pp} over the second-best method (DINO).

\subsection{Ablation Study}
\label{sec:ablation}
We conduct ablations on the Cholec80 phase recognition task by pretraining on LEMON~\cite{Che2025LEMON:Settings} and report top-1 accuracy under linear probing and \textit{k}-NN (\textit{k} = 20). 
We use a ViT-S/16 backbone to reduce computational cost.
Our default settings in the Tables~\ref{tab:ablation_component},~\ref{tab:ablation_loss_type}, and~\ref{tab:ablation_frames} are highlighted in \colorbox{Gray!20}{gray}.

\noindent\textbf{Effect of PL-Stitch Components.}
We analyze the contribution of each objective in Table~\ref{tab:ablation_component}. 
Our baseline (row 1), with only the $\mathcal{L}_{\text{MIM}}$ objective~\cite{Zhou2022IBOT:Tokenizer}, achieves 69.4\% \textit{k}-NN accuracy.
Adding our core temporal ranking objective ($\mathcal{L}_{\text{vid}}$) (row 2) provides the largest single boost, improving \textit{k}-NN accuracy by \textbf{9.5 pp} to 78.9\%. 
Adding only the spatio-temporal jigsaw objective ($\mathcal{L}_{\text{jigsaw}}$) (row 3) provides a smaller benefit. 
Finally, our full PL-Stitch model (row 4), combining all three objectives, achieves the highest accuracy at \textbf{80.2\%}. This demonstrates that our global temporal and local jigsaw objectives are complementary, and their joint optimization leads to the most robust representation.

\noindent
\textbf{Impact of PL Ranking Formulation.}
We compare our Plackett-Luce (PL) formulation against traditional ordering objectives (Table~\ref{tab:ablation_loss_type}). We replace our $\mathcal{L}_{\text{vid}}$ with two common baselines: 1) a Pairwise Loss~\cite{Hu2021ContrastLearning}, which enforces the order of all pairs in the sequence, and 2) a Permutation Classification loss~\cite{Misra2016ShuffleVerification}, which treats each shuffled permutation of the sequence as a distinct class. Both baselines perform worse than our PL objective, confirming that the probabilistic, listwise formulation of the PL model is key to learning a robust, progress-aware representation.

\noindent
\textbf{Number of Temporal Ranking Frames ($k$).}
We analyze the number of frames $k$ used in the temporal ranking objective ($\mathcal{L}_{\text{vid}}$) in Table~\ref{tab:ablation_frames}. 
Performance on the \textit{k}-NN evaluation improves by 2.1 pp when increasing $k$ from $4$ to $8$, as this provides a richer temporal context for ranking. 
While performance at $k = 12$ and $k=16$ is similar to $k = 8$ ($77.5\%$ vs $77.8\%$ on linear probing, $80.3\%$ vs $80.2\%$ on \textit{k}-NN), increasing the sequence length raises the computational burden (up to $4\times$ for $k = 16$). We therefore selected $k = 8$ as the optimal balance between accuracy and efficiency.

\subsection{Qualitative Results}
\label{sec:qualitative}

\noindent
\textbf{Temporal Consistency.}
As shown in Fig.~\ref{fig:progress_qualitative}, linear probe predictions from PL-Stitch on both AutoLaparo and GTEA are significantly more stable and temporally consistent than the competing methods, demonstrating PL-Stitch's ability to learn a robust procedural representation.

\noindent
\textbf{Feature Space Visualization.} 
Fig.~\ref{fig:tsne} visualizes the frozen Cholec80 features using t-SNE~\cite{VanDerMaaten2014AcceleratingAlgorithms}. Baseline features from VideoMAEv2, DINO, and iBOT  overlap heavily, showing no clear phase separation. In contrast, PL-Stitch forms distinct, well-separated clusters that align with ground-truth phases. 
Its clustering quality ($\text{ARI}: 0.3536, \text{NMI}: 0.4546$) is more than twice as high as the second-best baseline, DINO ($\text{ARI}: 0.1043, \text{NMI}: 0.2051$).
This clear separation visually explains its high \textit{k}-NN accuracy and confirms its learned procedural awareness.

\noindent 
\textbf{Attention Maps.} 
We choose the [CLS] token as the query and visualize its resulting attention map in Fig.~\ref{fig:attention}. The baselines show scattered and unstable attention, failing to track the surgical action. In contrast, our PL-Stitch maintains a stable and precise focus on the instrument and its operating area. This persistent localization provides qualitative evidence that our model learns the procedural context, not just static objects. Additional qualitative examples are in the supplementary material. \section{Conclusion}
\label{sec:conclusion}
We addressed the procedural agnosticism of modern self-supervised learning methods, first demonstrating with a motivating experiment that they are blind to procedural order. 
To solve this, we proposed PL-Stitch, a novel self-supervised framework that harnesses the video's inherent temporal order as a powerful supervisory signal. 
By integrating probabilistic Plackett-Luce ranking for both global workflow progression and fine-grained object correspondence, PL-Stitch sets a new state-of-the-art on challenging surgical and cooking benchmarks. 
This validates our core insight that explicitly modeling temporal order is key to learning procedurally-aware video representations. 
Future work will move beyond representation learning to generative tasks, such as action anticipation, and explore multi-modal integration by aligning the learned procedural steps with instructional text from recipes or surgical manuals.

\clearpage

\appendix
\clearpage
\setcounter{page}{1}
\maketitlesupplementary

\section{Method Details}
We provide the comprehensive mathematical formulation of our Plackett-Luce ranking framework and describe its implementation across the proposed video and image branches.

\subsection{Plackett-Luce Ranking Formulation}
Our framework leverages the \textbf{Plackett-Luce (PL) model}~\cite{Plackett1975ThePermutations, Luce1977TheYears} to structure both the global and local objectives as listwise ranking problems. The PL model is parameterized by a vector of positive real-valued scores $s = (s_1, \dots, s_K)$, where $K$ is the number of items being ranked. The score $s_i$ represents the overall utility or preference strength of item $i$.

The probability $P(r|s)$ of observing a specific full ranking (permutation) $r = (r(1), r(2), \dots, r(K))$, where $r$ is a reordering of $\{1, \dots, K\}$, is defined sequentially based on Luce's Choice Axiom~\cite{Luce1977TheYears}. Specifically, the probability is the product of the conditional probabilities of choosing item $r(i)$ from the set of items $R_i$ not yet ranked at step $i$:

\begin{equation}
\label{eq:pl}
    P(r|s)=\prod_{i=1}^{K}\frac{\exp(s_{r(i)})}{\sum_{j=i}^{K}\exp(s_{r(j)})}
\end{equation}

The model is trained by minimizing the negative log-likelihood of the single ground-truth permutation $r^* = (r^*(1), \dots, r^*(K))$.
The loss function $\mathcal{L}_{\text{PL}}$ is defined as:

\begin{align}
    \mathcal{L}_{\text{PL}}(s,r^{*}) = -\log P(r^{*}|s) \quad
    \label{eq:Lpl}
\end{align}

By applying the negative logarithm to the product in the PL definition, the loss $\mathcal{L}_{\text{PL}}$ unfolds into a sum of $K$ distinct preference decisions. This fully expanded form is minimized during training:

\begin{equation}
\label{eq:pl_loss_expanded}
\begin{split}
\mathcal{L}_{\text{PL}}(s,r^{*}) &= - \sum_{i=1}^{K} \left[ \log \left( \exp(s_{r^*(i)}) \right) \right. \\
&\quad - \left. \log \left( \sum_{j=i}^{K} \exp(s_{r^*(j)}) \right) \right]
\end{split}
\end{equation}

This simplifies to the final loss form:

\begin{equation}
\label{eq:pl_loss_simplified}
\mathcal{L}_{\text{PL}}(s,r^{*}) = \sum_{i=1}^{K} \left[   \log \left( \sum_{j=i}^{K} \exp(s_{r^*(j)}) \right) - s_{r^*(i)}\right]
\end{equation}

where $i$ indexes the rank position being determined, $r^*(i)$ denotes the item at rank $i$ in the ground-truth sequence $r^*$, and the inner summation over $j$ computes the total score of all items remaining in the unranked positions from $i$ to $K$.

\begin{figure}[!t]
    \centering
    \includegraphics[width=\linewidth]{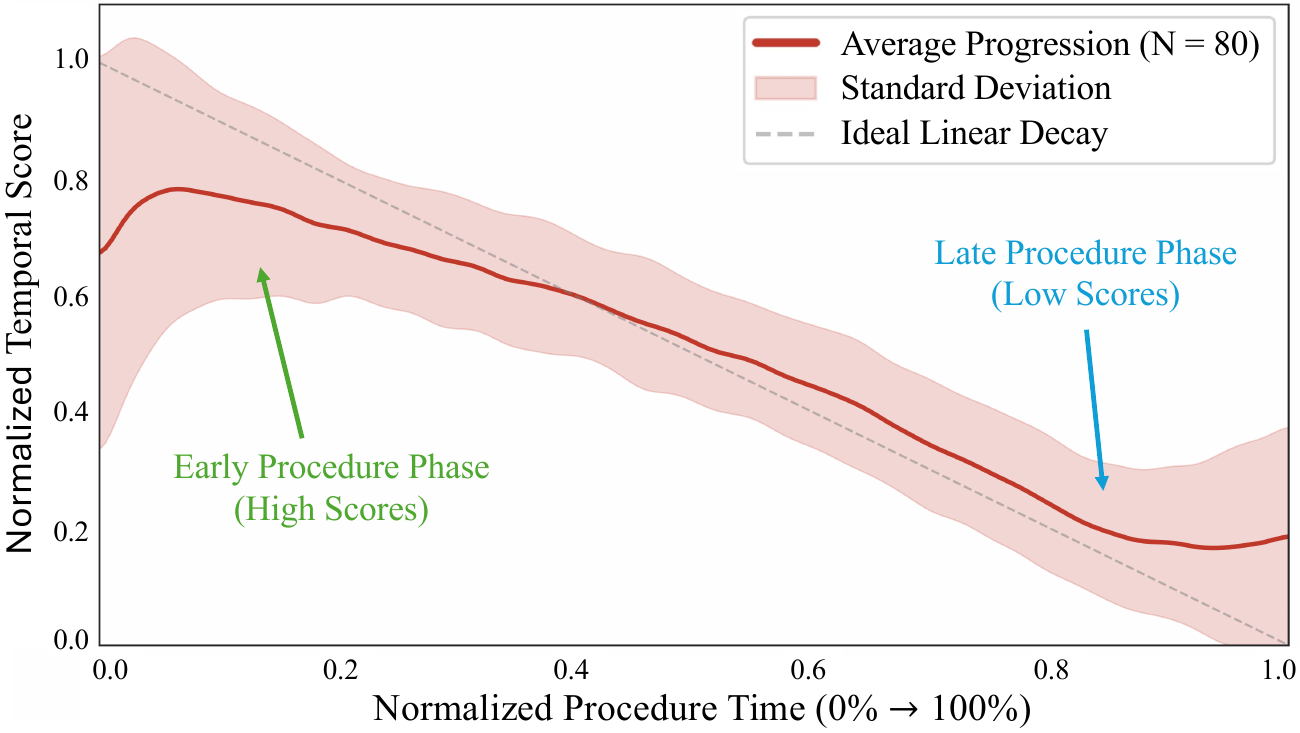}
    \caption{\textbf{Global Procedural Progression.} We visualize the temporal progression score averaged across all 80 Cholec80~\cite{Twinanda2017EndoNet:Videos} videos, where higher scores indicate earlier phases. To generate this plot, we extract [CLS] embeddings using the frozen backbone of our trained PL-Stitch model and predict frame-wise scores using its temporal head. These scores are normalized per video to a unit scale ($0 \to 1$) before computing the mean (solid line) and standard deviation (shaded region). Despite being trained on the different dataset, LEMON~\cite{Che2025LEMON:Settings}, the model successfully generalizes to the unseen Cholec80 videos. The predicted score consistently decreases during the active surgical workflow. The observed deviations at the boundaries ($t < 0.1$ and $t > 0.9$) correspond to the camera entering and exiting the body, effectively marking the non-operative transitions surrounding the procedure.}
    \label{fig:global_progression}
\end{figure}

This compels the encoder to assign a significantly higher score $s_{r^*(i)}$ to the correct frame (or patch) chosen at step $i$ compared to the log-sum-exponent of all remaining items. This probabilistic approach scales the error penalty proportionally to the ranking mistake severity, proving more robust than permutation classification~\cite{Misra2016ShuffleVerification, Wei2018LearningTime, Xu2019Self-SupervisedPrediction, Fernando2017Self-SupervisedNetworks} or pairwise losses~\cite{Hu2021ContrastLearning,Lee2017UnsupervisedSequences, Sermanet2017Time-ContrastiveObservation}.

\subsection{Video Branch: Listwise Temporal Ranking}
The Video Branch implements the global workflow progression objective $\mathcal{L}_{\text{vid}}$ as an instantiation of the $\mathcal{L}_{\text{PL}}$ loss.

\begin{itemize}
    \item \textbf{Items ($K$):} A \text{clip} $C_v=(v_1, \dots, v_K)$ of $K=8$ sampled frames is used. The PL parameters $s_{\text{\text{clip}}}=(s_1, \dots, s_K)$ are predicted from the [CLS] tokens of the frames via the temporal head $h_{\text{vid}}$.
    \item \textbf{Ground-Truth ($r^{*}$):} The target permutation is the true chronological order: $r_{\text{clip}}^{*} = (1, 2, \dots, K)$.
    \item \textbf{Loss Function:} The objective minimizes the difference between the predicted scores and the true temporal order:
    $$
    \mathcal{L}_{\text{vid}} = \mathcal{L}_{\text{PL}}(s_{\text{clip}}, r_{\text{clip}}^{*}).
    $$
\end{itemize}

\subsection{Image Branch: Spatio-temporal Jigsaw}
The Image Branch implements the $\mathcal{L}_{\text{\text{jigsaw}}}$ objective, another instantiation of the $\mathcal{L}_{\text{PL}}$ loss, designed to capture fine-grained object correspondence using local temporal context, inspired by ~\cite{Gupta2023SiameseAutoencoders, Liu2025WhenLearning}.

\begin{itemize}
    \item \textbf{Items ($N$):} The items are the $N$ patches of the central, masked frame $v_t$.

    \item \textbf{Temporal Context Injection:} Patches of the masked current frame serve as \textbf{Queries (Q)} ($\tilde{E}^{\text{patch}}$). The concatenated embeddings from adjacent \textbf{past} ($v_{t-\tau_1}$) and \textbf{future} ($v_{t+\tau_2}$) frames serve as \textbf{Keys (K) and Values (V)}, forcing the model to rely on temporal consistency for spatial reconstruction (see Sec.~\ref{sec:pretrain_details} for $\tau_1$ and $\tau_2$ sampling details).

    \item \textbf{Ground-Truth ($r^{*}$):} The target permutation is the original linearized patch order: $r_{\text{\text{jigsaw}}}^{*} = (1, 2, \dots, N)$.
    \item \textbf{Loss Function:} The \text{jigsaw} head $h_{\text{\text{jigsaw}}}$ predicts $s_{\text{\text{jigsaw}}}$, which is minimized using:
    $$
    \mathcal{L}_{\text{\text{jigsaw}}} = \mathcal{L}_{\text{PL}}(s_{\text{\text{jigsaw}}}, r_{\text{\text{jigsaw}}}^{*}). \quad 
    $$
\end{itemize}

\subsection{Learned Global Workflow Progression}
To investigate whether PL-Stitch captures the global procedural structure, we conducted a qualitative analysis on the Cholec80~\cite{Twinanda2017EndoNet:Videos} dataset. We employed the model pretrained on the large-scale LEMON dataset~\cite{Che2025LEMON:Settings}, which contains no samples from Cholec80, ensuring a strict zero-shot evaluation. Specifically, we processed all 80 downstream videos by extracting [CLS] embeddings via the \textbf{frozen PL-Stitch backbone} and subsequently generating frame-wise real-valued scores using the \textbf{pretrained temporal head}.

Fig.~\ref{fig:global_progression} illustrates the temporal progression score averaged across all videos, with the procedural duration normalized to a unit scale ($0 \to 1$). Despite being pretrained with an 8-frame ranking objective, the model exhibits a remarkable capacity to generalize globally. The average score demonstrates a consistently decreasing trend over the course of the procedure. This behavior is a direct consequence of the Plackett-Luce optimization: to maximize the likelihood of the correct chronological order, the model is trained to assign larger scores to the earlier frames in any sampled sequence. This confirms that PL-Stitch has effectively learned to map the visual evolution of the procedure to a continuous scalar representation of procedural progress.

Notably, we observe a slight deviation from strict monotonicity at the extreme temporal boundaries (approx. $t < 0.1$ and $t > 0.9$). This behavior aligns with the visual semantics of the Cholec80 dataset. The initial ``Preparation'' and final ``Retraction'' phases share similar visual characteristics, as both involve the camera entering or exiting the body. These frames often depict blurry views of the abdominal wall or out-of-body scenes where no surgical tools are present. Consequently, the model assigns the peak ``earliness'' score not to these ambiguous pre-operative frames, but to the onset of the first active surgical phase ($t \approx 0.1$). This suggests that PL-Stitch goes beyond simple frame counting. It identifies the effective start of the operative workflow by recognizing the visual cues of active surgery while distinguishing them from non-informative idle states at the video boundaries.

\section{Implementation Details}
We provide the implementation specifics and hyperparameters for our pretraining framework.
\subsection{Pretraining Details}
\label{sec:pretrain_details}
We detail the implementation settings for our pretraining objective, which is the weighted sum $\mathcal{L}_{\text{total}}=\lambda_{1}\mathcal{L}_{\text{vid}}+\lambda_{2}\mathcal{L}_{\text{MIM}}+\lambda_{3}\mathcal{L}_{\text{\text{jigsaw}}}$.

\begin{itemize}
    \item \textbf{Backbone and Optimizer:} We use a ViT-B/16~\cite{Dosovitskiy2020AnScale} backbone. The optimizer is AdamW, with a base learning rate of $4 \times 10^{-4}$, a weight decay of $0.05$, and a total batch size of 240.

    \item \textbf{Pretraining Length:} Surgical models are pretrained on the LEMON dataset~\cite{Che2025LEMON:Settings} for 30 epochs. Cooking models are pretrained directly on their respective official training sets for 100 epochs. These durations were specifically selected to guarantee that the models are trained sufficiently well and that the training loss has effectively converged for each domain.

    \item \textbf{Image-branch dataset construction:} We sample the pretraining videos at a rate of 1 fps to construct the dataset for the image branch.

    \item \textbf{Video-branch dataset construction:}
    We extract 8-frame clips to construct the dataset for the video branch. We sample multiple clips from each video, scaling the count with the video duration, to ensure the final size of the clip dataset is identical to that of the image branch.

    \item \textbf{Maximum iterations $L$}: This is calculated by dividing the total image-branch dataset size by the batch size, representing the number of steps required to complete one full epoch.

    \item \textbf{$\mathcal{L}_{\text{vid}}$ Parameters:} The \text{clip} length is set to $K=8$ frames.

    \item \textbf{$\mathcal{L}_{\text{MIM}}$ Parameters:} A block-wise masking ratio of $30\%$ is applied to the current frame $v_t$, following the iBOT protocol~\cite{Zhou2022IBOT:Tokenizer}.

    \item \textbf{$\mathcal{L}_{\text{jigsaw}}$ parameters}: Following~\cite{Liu2025WhenLearning}, which samples context frames from temporal offsets in the range $\pm[0.15T, 0.25T]$ on Kinetics-400~\cite{Carreira2017QuoDataset} videos (average duration $T$ $\approx 10$\,s), we randomly sample the temporal offsets $\tau_1$ and $\tau_2$ from the range $[1.5, 2.5]$\,s relative to the current frame.

\end{itemize}

\begin{table}[!t]
    \centering
    \small
    \setlength{\tabcolsep}{4.7pt} \caption{
        \textbf{Ablation on loss weights ($\lambda$)}.
    }
    \label{tab:ablation_lambda}
\begin{tabularx}{.91\linewidth}{@{}lccccc@{}}
    \hline
        No. & $\lambda_1$ ($\mathcal{L}_{\text{vid}}$) & $\lambda_2$ ($\mathcal{L}_{\text{MIM}}$) & $\lambda_3$ ($\mathcal{L}_{\text{\text{jigsaw}}}$) & Linear & \textit{k}-NN \\
    \hline
        1 & $0.0$ & $1.0$ & $0.0$ & $73.4$ & $69.4$ \\
        2 & $1.0$ & $1.0$ & $0.0$ & $77.1$ & $78.9$ \\
        3 & $1.0$ & $1.0$ & $1.0$ & $76.9$ & $78.4$ \\
        4 & $1.0$\cellcolor{Gray!20} & $1.0$\cellcolor{Gray!20} & $0.4$\cellcolor{Gray!20} & $\mathbf{77.8}$\cellcolor{Gray!20} & $\mathbf{80.2}$\cellcolor{Gray!20} \\
        5 & $0.5$ & $1.0$ & $0.4$ & $77.0$ & $78.5$ \\
        6 & $2.0$ & $1.0$ & $0.4$ & $77.4$ & $79.5$ \\
    \hline
    \end{tabularx}
\end{table}

\subsection{Ablation on Loss Weights}
We investigate the contribution of each objective and the sensitivity to loss weights on the Cholec80 phase recognition task. Results are reported in Table~\ref{tab:ablation_lambda}, where our optimal configuration is highlighted in \colorbox{Gray!20}{gray}.
First, the addition of the temporal ranking loss $\mathcal{L}_{\text{vid}}$ (row 2) yields a substantial gain over the MIM-only baseline (row 1), boosting \textit{k}-NN accuracy by \textbf{+9.5 pp} (from 69.4\% to 78.9\%). This confirms that explicit temporal ordering is the primary driver of procedural awareness.

Regarding the jigsaw weight $\lambda_3$, we observe that while removing it entirely (row 2) lowers accuracy, increasing it to $\lambda_3=1.0$ (row 3) also proves suboptimal ($78.4\%$ in \textit{k}-NN). This suggests that while local spatio-temporal context is beneficial, an excessive jigsaw weight may distract the model from learning the global workflow progression.
Finally, regarding the main temporal weight $\lambda_1$, our default value of 1.0 (row 4) outperforms both halving (row 5) and doubling (row 6) the weight, achieving the peak \textit{k}-NN performance of \textbf{$80.2\%$}.

\section{Downstream Task Details}
We evaluate feature quality on five procedural video benchmarks for temporal phase recognition and action segmentation.

\subsection{Evaluation Datasets}
We evaluate our method on five challenging procedural benchmarks, covering both the surgical and cooking domains.

\begin{itemize}
    \item \textbf{Cholec80~\cite{Twinanda2017EndoNet:Videos}}: This dataset consists of 80 videos of laparoscopic cholecystectomy procedures. The objective is surgical phase recognition, a frame-wise classification task where the model must assign one of 7 distinct surgical phases (e.g., Preparation, Calot-Triangle Dissection, Clipping and Cutting) to every frame in the video.

    \item \textbf{AutoLaparo~\cite{Wang2022AutoLaparo:Hysterectomy}}: A dataset containing 21 videos of laparoscopic hysterectomy procedures. The task is surgical phase recognition, challenging the model to identify the current phase from 7 defined surgical phases across long, untrimmed videos.

    \item \textbf{M2CAI16~\cite{Stauder2016TheChallenge}}: This dataset features 41 videos of laparoscopic cholecystectomy. 
The task is surgical phase recognition, where the model must recognize 8 surgical phases.

    \item \textbf{Breakfast~\cite{Kuehne2014TheActivities}}: A dataset comprising 1712 videos capturing 10 different breakfast preparation activities (e.g., making coffee, pancakes) from a third-person perspective. The task is temporal action segmentation, which involves densely classifying video frames into 48 fine-grained action steps (e.g., ``pour milk'', ``crack egg'').

    \item \textbf{GTEA~\cite{Fathi2011LearningActivities}}: An egocentric (first-person) dataset containing 28 videos of daily cooking activities. The task is temporal action segmentation across 11 action classes, presenting unique challenges due to severe camera motion and hand occlusions typical of wearable cameras.
\end{itemize}

\subsection{Evaluation Metrics}
\textbf{Linear probing.}
We evaluate the quality of the learned representations using the following standard metrics under the linear probing protocol.

\begin{itemize}
    \item \textbf{Accuracy (Acc).} For all tasks, we report standard frame-wise top-1 accuracy. It is computed as the ratio of correctly classified frames to the total number of frames $T$:
    \[
        \text{Acc} = \frac{1}{T} \sum_{t=1}^{T} \mathbb{I}(\hat{y}_t = y_t),
    \]
    where $\hat{y}_t$ is the predicted class for frame $t$, $y_t$ is the ground-truth class, and $\mathbb{I}(\cdot)$ is the indicator function.

    \item \textbf{F1-score (F1).} For surgical phase recognition, we report the frame-wise macro F1-score, defined as the average of the per-class F1-scores. For each class $c$, the F1-score is the harmonic mean of precision and recall:
    \[
        \text{F1}_c = 2 \cdot \frac{\text{Precision}_c \cdot \text{Recall}_c}{\text{Precision}_c + \text{Recall}_c},
    \]
    with
    \[
        \text{Precision}_c = \frac{TP_c}{TP_c + FP_c}, \quad
        \text{Recall}_c = \frac{TP_c}{TP_c + FN_c},
    \]
    where $TP_c$, $FP_c$, and $FN_c$ denote the numbers of true positives, false positives, and false negatives for class $c$, respectively. The reported macro F1-score is then
    \[
        \text{F1}_{\text{macro}} = \frac{1}{C} \sum_{c=1}^{C} \text{F1}_c,
    \]
    where $C$ is the total number of classes.

    \item \textbf{Segmental F1@$\delta$.} For cooking action segmentation, we evaluate the quality of predicted segments using the segmental F1-score at Intersection over Union (IoU) thresholds $\delta \in \{10\%, 25\%, 50\%\}$. A predicted segment is counted as a true positive ($TP@\delta$) if its IoU with a ground-truth segment of the same class exceeds $\delta$ (and each ground-truth segment is matched to at most one prediction). The segmental F1@$\delta$ is defined as
    \[
        \text{F1}@\delta = 2 \cdot \frac{\text{Precision}@\delta \cdot \text{Recall}@\delta}{\text{Precision}@\delta + \text{Recall}@\delta},
    \]
    where
    \[
        \text{Precision}@\delta = \frac{TP@\delta}{N_{\text{pred}}}, \quad
        \text{Recall}@\delta = \frac{TP@\delta}{N_{\text{gt}}},
    \]
    and $N_{\text{pred}}$ and $N_{\text{gt}}$ are the numbers of predicted and ground-truth segments, respectively.

    \item \textbf{Edit distance (Edit).} For cooking tasks, we also measure temporal ordering consistency using a normalized Levenshtein edit score. Let $S_{\text{pred}}$ and $S_{\text{gt}}$ denote the predicted and ground-truth sequences of action segments, respectively. The edit score is defined as
    \[
        \text{Edit} = \left( 1 - \frac{\text{lev}(S_{\text{pred}}, S_{\text{gt}})}{\max(|S_{\text{pred}}|, |S_{\text{gt}}|)} \right) \times 100,
    \]
    where $\text{lev}(\cdot, \cdot)$ is the Levenshtein distance between two sequences and $|\cdot|$ denotes sequence length (number of segments).
\end{itemize}

\begin{table}[!t]
    \centering
    \small
    \caption{
    \textbf{Five-fold cross-validation results on surgical datasets.} We report the mean $\pm$ std of the linear probing accuracy.
    }
    \label{tab:cross_val_surgical}
    \begin{tabularx}{\linewidth}{@{}l cccc@{}}
        \toprule
        \textbf{Method} & \textbf{AutoLaparo} & \textbf{Cholec80} & \textbf{M2CAI16} \\
        \midrule
        VideoMAEv2~\cite{Wang2023VideoMAEMasking} & $50.3 \pm 1.9$ & $56.5 \pm 1.7$ & $51.4 \pm 3.4$ \\
        DINO~\cite{Caron2021EmergingTransformers} & $75.5 \pm 1.6$ & $73.5 \pm 1.5$ & $68.9 \pm 2.3$ \\
        iBOT~\cite{Zhou2022IBOT:Tokenizer} & $75.6 \pm 2.2$ & $75.9 \pm 1.2$ & $71.5 \pm 2.1$ \\
        \textbf{PL-Stitch (Ours)} & $\mathbf{80.1 \pm 2.5}$ & $\mathbf{82.6 \pm 1.8}$ & $\mathbf{75.2 \pm 2.0}$ \\
        \bottomrule
    \end{tabularx}
\end{table}

\begin{figure}[!t] 
    \centering
\begin{subfigure}[b]{\linewidth}
        \centering
        \includegraphics[width=\linewidth]{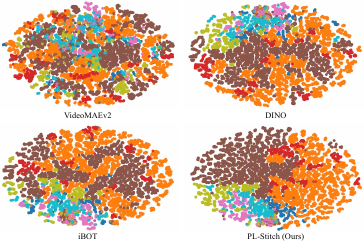} 
        \caption{Perplexity = 10 (Local Structure)}
        \label{fig:tsne_ple10}
    \end{subfigure}
    
    \vspace{0.2cm} 

\begin{subfigure}[b]{\linewidth}
        \centering
        \includegraphics[width=\linewidth]{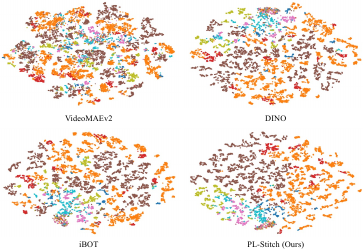} \caption{Perplexity = 30 (Balanced / Default)}
        \label{fig:tsne_ple30}
    \end{subfigure}

    \vspace{0.2cm}

\begin{subfigure}[b]{\linewidth}
        \centering
        \includegraphics[width=\linewidth]{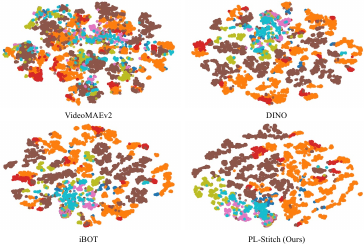}
        \caption{Perplexity = 50 (Global Structure)}
        \label{fig:tsne_ple50}
    \end{subfigure}
    
    \caption{\textbf{Robustness of feature embeddings on the Cholec80 dataset under varying t-SNE parameters.} Comparison of feature visualizations at (a) Perplexity 10, (b) the default Perplexity 30, and (c) Perplexity 50. While baselines show mixed clusters across all settings, PL-Stitch consistently maintains clearer class separation.}
    \label{fig:tsne_robustness}
\end{figure}

\begin{figure*}[!t]
    \centering
    \includegraphics[width=.9\linewidth]{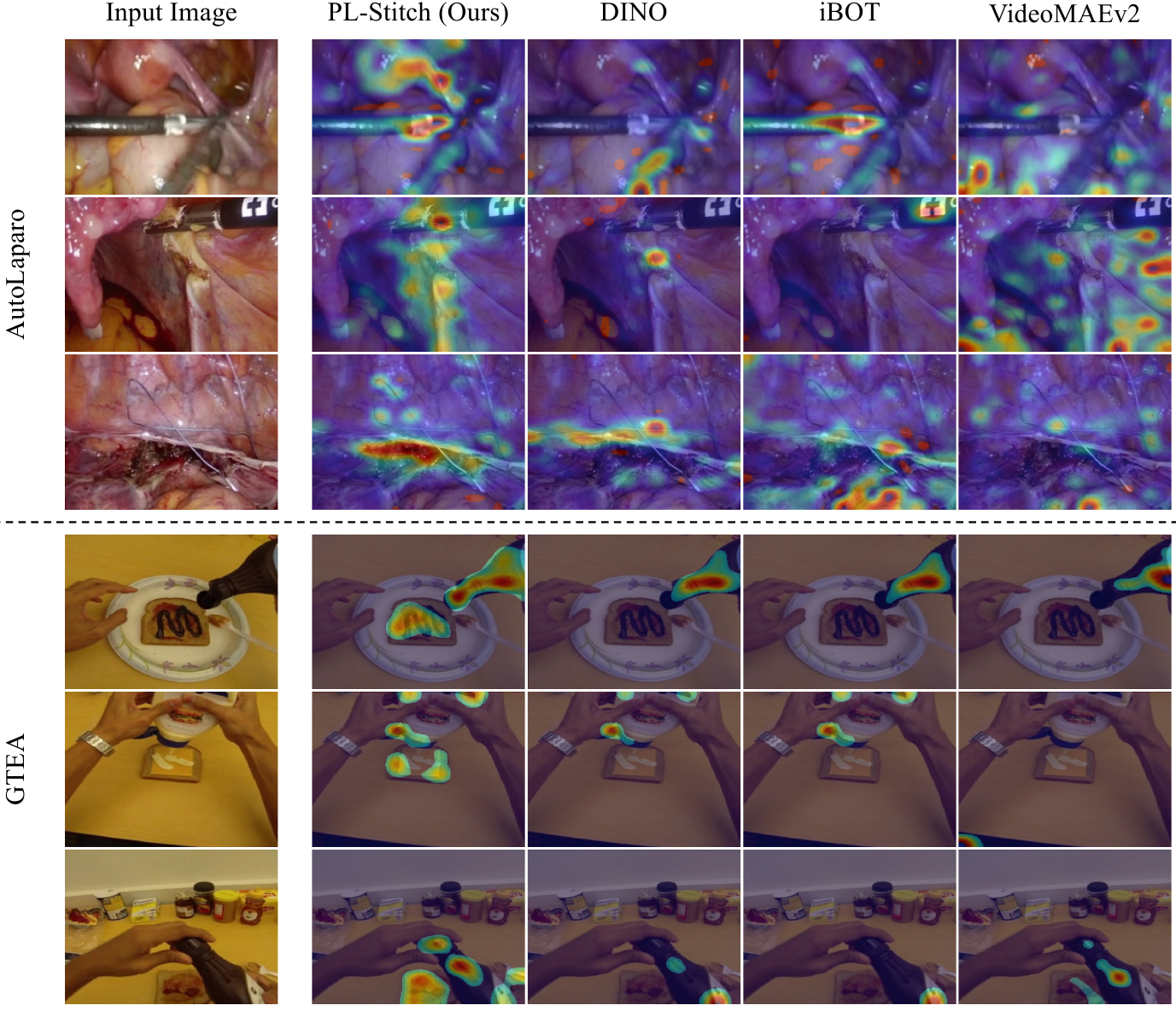} 
    \caption{
        \textbf{Qualitative comparison of attention localization across diverse procedural scenes.}
        We visualize the attention maps queried by the [CLS] token for our method (PL-Stitch) and other models (DINO, iBOT, VideoMAEv2) on input images from AutoLaparo (top) and GTEA (bottom). PL-Stitch consistently demonstrates a superior ability to localize and focus its attention on key interaction areas, such as surgical instruments or manipulated objects. This outperforms other methods that exhibit more diffuse or misplaced attention.
        }
    \label{fig:more_attn}
\end{figure*}

\noindent
\textbf{\textit{k}-NN evaluation.}
Following established self-supervised learning benchmarks~\cite{Caron2021EmergingTransformers, Zhou2022IBOT:Tokenizer, Oquab2023DINOv2:Supervision}, we assess the quality of the frozen feature space using non-parametric $k$-Nearest Neighbor ($k$-NN) classification. We first extract features from the frozen backbone for all frames in the training and testing sets. For each test frame, we retrieve its $k=20$ nearest neighbors from the training set based on cosine similarity. The predicted class is determined via weighted majority voting, where neighbors are weighted by their similarity scores. This protocol provides a direct measure of the semantic separability of the learned representation without requiring parameter updates. We report the top-1 accuracy.

\section{Additional Experimental Results}
We present further quantitative and qualitative analyses to demonstrate the statistical robustness and semantic interpretability of the representations learned by PL-Stitch.
\subsection{Five-Fold Cross-Validation on Surgical Phase Recognition}
In the surgical domain, robust evaluation is critical to ensure that models generalize effectively across varying patient anatomies and surgical workflows. To verify the statistical robustness of our method in this challenging setting, we performed 5-fold cross-validation across all three surgical datasets: Cholec80~\cite{Twinanda2017EndoNet:Videos}, AutoLaparo~\cite{Wang2022AutoLaparo:Hysterectomy}, and M2CAI16~\cite{Stauder2016TheChallenge}. We report the linear probing top-1 accuracy (Mean $\pm$ Std) to demonstrate the stability of the learned features.

As shown in Table~\ref{tab:cross_val_surgical}, PL-Stitch yields the highest mean accuracy across all datasets while maintaining a low standard deviation. This confirms that the procedurally-aware representations learned by our model are not only discriminative but also highly stable across different data splits and surgical domains.

\begin{figure*}[!t]
    \centering
    \includegraphics[width=.84\linewidth]{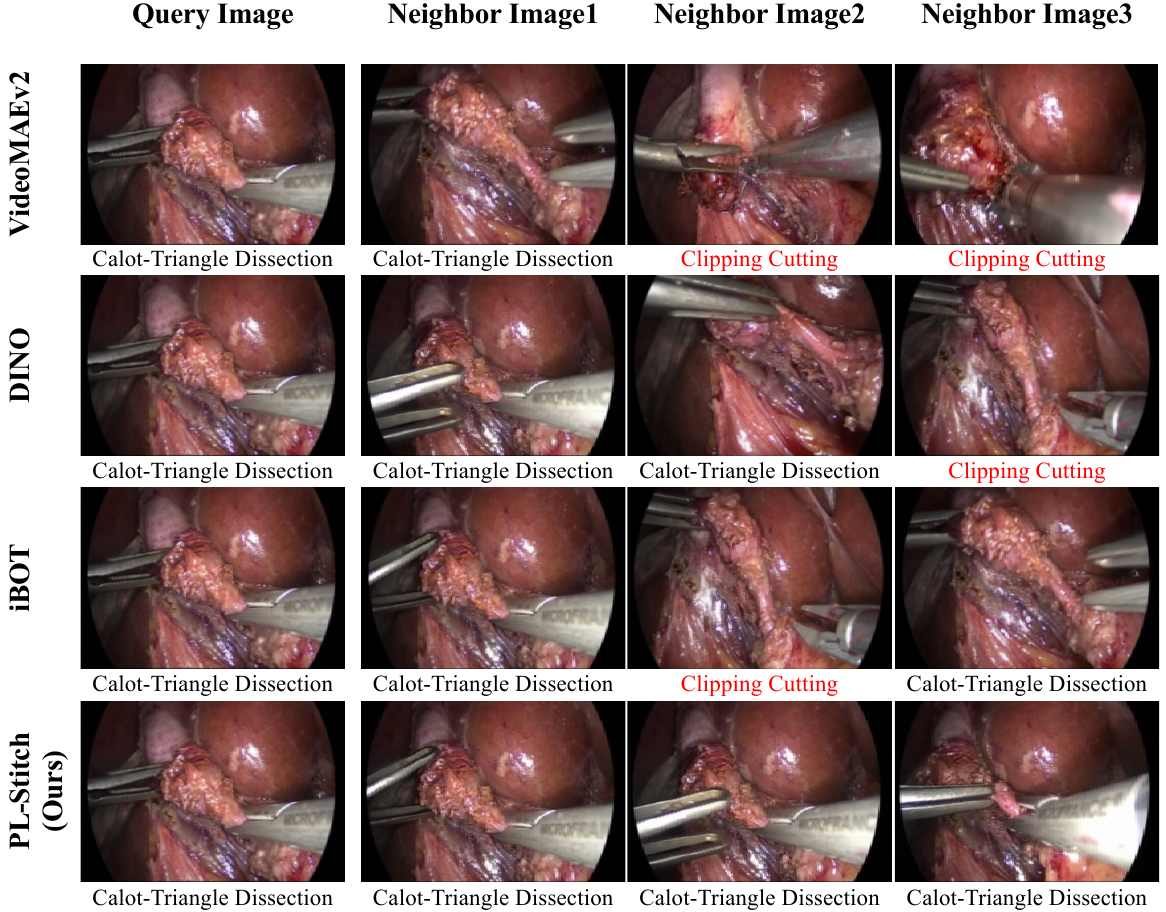} 
    \caption{
    \textbf{Nearest Neighbor Retrieval.}
    Comparison of the top-3 retrieved frames for a query image from the Calot-Triangle Dissection phase. Incorrect phase predictions are highlighted in red text.
Baselines such as VideoMAEv2, DINO, and iBOT are deceived by visual similarity and incorrectly retrieve frames from the Clipping Cutting phase. PL-Stitch retrieves only procedurally synchronous frames, unlike baselines which fail to distinguish between similar-looking but distinct procedural steps.}
    \label{fig:knn_qualitative}
\end{figure*}

\subsection{Sensitivity Analysis of t-SNE Visualization}
In the qualitative evaluation presented in the main manuscript, we employed a default t-SNE perplexity of 30, which offers a balanced representation of both local and global feature structures. To verify that the observed class separability is an intrinsic property of the learned embeddings rather than an artifact of visualization parameter tuning, we provide a robustness analysis on the Cholec80 dataset~\cite{Twinanda2017EndoNet:Videos} in Fig.~\ref{fig:tsne_robustness}.

We visualize feature embeddings at perplexities 10, 30, and 50 (Figs.~\ref{fig:tsne_ple10}, \ref{fig:tsne_ple30}, \ref{fig:tsne_ple50}). At perplexity 10, the focus on local neighborhoods causes fragmentation, yet PL-Stitch retains identifiable groupings. Increasing the perplexity to 30 and 50 reveals distinct and well-separated clusters for our method as global structure becomes emphasized. Conversely, VideoMAEv2, DINO, and iBOT show persistent overlap between similar phases across all settings. This consistency confirms the robustness of our learned feature space.

\subsection{Additional Attention Maps}
We provide an extended qualitative analysis of the attention focus of the model by visualizing self-attention maps queried by the [CLS] token across diverse procedural contexts in Fig.~\ref{fig:more_attn}. This comparison encompasses both surgical scenes from the AutoLaparo dataset~\cite{Wang2022AutoLaparo:Hysterectomy} and cooking activities from the GTEA dataset~\cite{Fathi2011LearningActivities}. PL-Stitch consistently concentrates high attention weights within task-relevant areas and demonstrates a strong semantic alignment with the workflow. 
For instance, in the AutoLaparo examples, our PL-Stitch model's attention remains anchored on the instrument-tissue interaction sites and demonstrates robust tracking of the surgical flow. Similarly, in the GTEA examples, attention accurately tracks the manipulated objects and active interaction zones, such as the condiment container and the spread on the bread.
In contrast, baseline methods such as DINO, iBOT, and VideoMAEv2 exhibit significantly more diffuse attention patterns that often drift towards background elements or fail to distinctly highlight the active interaction site. This comparison underscores the stability and precision of PL-Stitch in localizing key visual cues compared to prior self-supervised approaches.

\subsection{Semantic Feature Retrieval}
Fig.~\ref{fig:knn_qualitative} shows a nearest-neighbor retrieval comparison on the Cholec80 dataset. Given a query image, baseline models frequently retrieve images that appear visually similar but belong to the wrong procedural phase. In contrast, PL-Stitch correctly retrieves images only from the correct phase and demonstrates a robust understanding of the underlying procedural workflow.

\FloatBarrier

{
    \small
    \bibliographystyle{ieeenat_fullname}
    \bibliography{main}
}

\end{document}